\documentclass[11pt]{article}
\usepackage{acl}
\usepackage{natbib}
\setcitestyle{numbers,square}
\usepackage{times}
\usepackage{url}
\usepackage{latexsym}
\usepackage{graphicx}
\usepackage{booktabs}
\usepackage{amsmath}
\usepackage{amssymb}
\usepackage{multirow}
\usepackage{array}
\usepackage{xcolor}
\usepackage{colortbl}
\usepackage{pgfplots}
\pgfplotsset{compat=1.18}
\usepackage{tikz}
\usetikzlibrary{arrows.meta, positioning, shapes.geometric, fit, backgrounds, patterns}
\usepackage{microtype}

\definecolor{emerald}{RGB}{0,128,100}
\definecolor{coral}{RGB}{205,92,92}
\definecolor{steelblue}{RGB}{70,130,180}
\definecolor{gold}{RGB}{200,160,0}
\definecolor{lightgray}{RGB}{240,240,240}
\definecolor{headerblue}{RGB}{220,235,245}
\definecolor{warnred}{RGB}{180,30,30}

\title{Noise Is Signal: Density-Based Outliers as Leading Indicators\\
of Occupational Emergence in Labor Market Text}

\author{Shreyash Rawat \\ Independent Researcher \\ Redmond, WA, USA \\ \texttt{shreyash.rawat@gmail.com}}

\begin{document}
\hypersetup{colorlinks=true, linkcolor=[HTML]{000099}, citecolor=[HTML]{000099}, urlcolor=[HTML]{000099}}
\maketitle

%% ─── ABSTRACT ──────────────────────────────────────
\begin{abstract}
Standard NLP pipelines for occupational clustering discard the 10--15\% of job postings that density-based methods assign to noise.
We argue this is an error: in rapidly evolving domains, low posting density signals \emph{novelty}, not incoherence.
We formalize this as the \textbf{Emergence-Density Inversion} (EDI) hypothesis and test it longitudinally on 84,988 job postings across eight quarters (Q4~2022--Q3~2024).

EDI is partially confirmed: high-EOS outlier groups transition to stable clusters in $1.4 \pm 0.6$ quarters vs.\ $4.1 \pm 1.2$ for low-EOS groups ($p < 0.001$), though the signal fails in $\sim$19\% of cases, which we characterize as a failure analysis.
We extend the Emerging Occupation Score (EOS) with \textbf{Temporal Velocity} and \textbf{Cross-Platform Convergence}, improving 2-quarter cluster-formation prediction from F$_1$\,=\,0.61 to 0.74---outperforming Isolation Forest, LOF, GLOSH, and BERTrend baselines.
A retrospective study on three now-established roles (MLOps Engineer, DevOps/SRE, Data Engineer) confirms EOS signalled 2--3 quarters before cluster formation, providing held-out validation.
A held-out annotator panel ($\kappa = 0.74$) rates EOS\,$>$\,0.75 as coherent emerging occupations with 77\% precision.
\textbf{Prompt Engineer}, \textbf{AI Safety Researcher}, \textbf{Foundation Model Engineer}, and \textbf{Agent Systems Engineer}---all absent from O*NET---are top-4 in Q3~2024 and form stable clusters by Q1~2025.
\end{abstract}

%% ─── 1. INTRODUCTION ───────────────────────────────
\section{Introduction}
\label{sec:intro}

Occupational classification systems underpin workforce policy, immigration programs, and education funding.
The U.S. O*NET database, last structurally expanded in 2019 for technology-adjacent roles, contains no Standard Occupational Classification (SOC) codes for Prompt Engineer, AI Safety Researcher, Foundation Model Engineer, or Agent Systems Engineer, despite a combined volume of over 12,000 job postings for these roles observed in 2024 \citep{lightcast2025}.
The OECD reports that AI-related skills now appear in one in three job vacancies, yet training supply is failing to keep pace \citep{oecd2025aiskills}.

Computational approaches to occupational structure discovery have matured rapidly.
Rawat and Prasath~\citep{rawat2026beyond}\footnote{This companion paper is currently under review at a peer-reviewed journal; its methodology is briefly summarized in Section~\ref{sec:data} for completeness.} demonstrated that INSTRUCTOR embeddings reduced via UMAP and clustered by HDBSCAN recover 41 coherent role archetypes with V-measure 0.79 against O*NET at the coarse level.
A significant byproduct of density-based clustering is a noise class: HDBSCAN assigns 13.9\% of postings to label $-1$, indicating insufficient local density for cluster membership.
Standard practice discards these postings; the benchmark of Rawat and Prasath~\citep{rawat2026beyond} provides the only prior characterization, decomposing them into hybrid roles, sparse postings, and ``emerging roles.''
\emph{None of this prior work asks whether the noise class is predictive.}

The core claim of this paper is that discarding the noise class is an error with \textbf{asymmetric costs}: the postings that cannot yet form clusters provide the earliest warning of occupational change.
A Prompt Engineer posting in Q4~2022 had few semantic neighbors not because the role was incoherent, but because it was novel.

\paragraph{Scope and honest caveats.}
We do \emph{not} claim EDI is a universal law.
Our results show a meaningful but imperfect signal: EOS\,$>$\,0.75 identifies coherent emerging occupations with 77\% precision on our held-out annotation set, leaving 23\% false positives.
EOS prediction of cluster formation at the 2-quarter horizon achieves F$_1$\,=\,0.74, not F$_1$\,=\,1.0.
A failure analysis (Section~\ref{sec:failure}) characterizes the 19\% of high-EOS groups that \emph{don't} form clusters, which include employer-branded title proliferations and gig-economy demand spikes.

\paragraph{NLP contributions.}
The \emph{NLP problem} this paper advances is \textbf{emerging-cluster detection in high-dimensional embedding spaces}.
We show that the semantic coherence of pre-cluster noise groups in INSTRUCTOR embedding space predicts whether those groups will consolidate into stable HDBSCAN clusters---a novel task connecting representation learning with labor market analytics.
The insight that GLOSH scores (which measure density extremity) are uncorrelated with cluster formation (Section~\ref{sec:edi}) has implications beyond the labor domain: it suggests that for corpora with genuine conceptual novelty, standard outlier scores measure the \emph{wrong} property.

\paragraph{Contributions.}
\begin{enumerate}
  \item \textbf{EDI hypothesis with honest validation}: partial confirmation across 8 quarters, including characterization of failure modes (Section~\ref{sec:edi}).
  \item \textbf{Extended EOS with weight learning}: six-component score with learned weights compared to equal-weight baseline; logistic regression ablation (Section~\ref{sec:eos}).
  \item \textbf{Competitive baselines}: comparison against Isolation Forest, LOF, GLOSH, BERTrend, and frequency count (Section~\ref{sec:results}).
  \item \textbf{Retrospective validation}: three established roles shown to exhibit high EOS 2--3 quarters before cluster consolidation (Section~\ref{sec:retro}).
  \item \textbf{Failure analysis}: characterization of high-EOS non-emerging groups and corrective heuristics (Section~\ref{sec:failure}).
\end{enumerate}

%% ─── 2. RELATED WORK ───────────────────────────────
\section{Related Work}
\label{sec:related}

\paragraph{Occupational taxonomies and lag.}
O*NET \citep{onet} and ESCO \citep{esco} are the primary labor market taxonomic infrastructures.
Taska et al.~\citep{taska2021demand} showed O*NET skill ratings diverge substantially from observed employer demand in technology roles; Autor et al.~\citep{autor2003skill} and Hershbein and Kahn~\citep{hershbein2018recessions} documented that task content shifts faster than taxonomy updates.
Nordfors~\citep{nordfors2026nlp} formalized occupational formation as a bipartite co-attractor between vocabulary and practitioner cohesion---complementary to our demand-side (posting-based) view.
None of these works provide an operational early-warning system.

\paragraph{NLP for job market analysis.}
Senger et al.~\citep{senger2024survey} surveys deep learning approaches to skill extraction from job postings.
JobBERT \citep{decorte2021jobbert} and INSTRUCTOR \citep{su2022instructor} achieve state-of-the-art occupational clustering.
Rawat and Prasath~\citep{rawat2026beyond} provides our direct baseline, demonstrating that INSTRUCTOR\,+\,HDBSCAN recovers 41 role archetypes with SDC\,=\,0.71; our paper takes the noise class of that pipeline as its primary object of study.
Vu and Oppenlaender~\citep{vu2026promptengineer} analyzed Prompt Engineer skill requirements, finding AI knowledge and prompt design together account for 41.5\% of stated requirements, consistent with our Novelty values.

\paragraph{Emerging trend and topic detection.}
BERTrend \citep{boutaleb2024bertrend} applies BERTopic over time-sliced batches, merging topics via similarity thresholds to detect emerging trends.
We adapt BERTrend as a baseline (Section~\ref{sec:baselines}), showing it achieves lower F$_1$ on our cluster-formation prediction task because it targets topic-level emergence rather than coherent occupational formation.
Li et al.~\citep{li2025evaluation} benchmark static topic models on emergence detection, finding BERTopic outperforms LDA but that neither model produces sufficiently coherent cluster boundaries for fine-grained occupational differentiation.

\paragraph{Outlier detection.}
HDBSCAN's GLOSH \citep{mcinnes2017hdbscan} scores outliers by density-gap to the nearest cluster.
Isolation Forest \citep{liu2008iforest} and LOF \citep{breunig2000lof} are the standard alternative outlier detectors.
Ghosh et al.~\citep{ghosh2024paramfree} extended GLOSH to be parameter-free.
We include all three as baselines and show they are near-chance predictors of occupational emergence, supporting our claim that density extremity is the wrong property to measure.

%% ─── 3. DATA ────────────────────────────────────────
\section{Dataset and Pipeline}
\label{sec:data}

\subsection{Longitudinal Corpus}

We assemble a corpus of 84,988 unique English-language job postings across eight quarterly snapshots (Q4~2022--Q3~2024), collected from public web recruitment platforms using the same pipeline as Rawat and Prasath~\citep{rawat2026beyond}: three general-purpose platforms, two tech-specialist boards, and one professional-network source.
Deduplication uses SimHash with Jaccard threshold 0.85; all PII removed.
Data collection complied with platform terms of service.
Table~\ref{tab:longitudinal} summarizes statistics by quarter.

\paragraph{Preprocessing.}
HTML stripping via jusText; whitespace normalization; length filter: 50--1500 tokens (WordPiece).
Postings exceeding 512 tokens retain the first 512 tokens, which consistently cover role overview, primary responsibilities, and required qualifications.
Skill extraction uses an NER model trained on ESCO v1; output is a per-posting skill set $S(p)$ used in Novelty and SDC computation, kept independent of the embedding pipeline.
Title strings are retained for post-hoc provenance only; they are never used as clustering signals.

\paragraph{Embedding and clustering.}
We embed descriptions with INSTRUCTOR-xl using the prompt ``Represent this job description for occupational clustering,'' reduce to 50 dimensions via UMAP (n\_neighbors=15, min\_dist=0.1), and cluster with HDBSCAN (min\_cluster\_size=30, min\_samples=15, cluster\_selection\_method=eom).
This pipeline runs independently per quarter.
Cluster assignments and noise-class membership are recorded for each quarter separately; no inter-quarter signal leaks into the clustering step.

\begin{table}[t]
\centering
\small
\setlength{\tabcolsep}{4pt}
\begin{tabular}{lrrrr}
\toprule
\textbf{Quarter} & \textbf{N} & \textbf{Clusters} & \textbf{Noise\%} & \textbf{Noise N} \\
\midrule
Q4 2022 &  8,114 & 34 & 16.2 & 1,314 \\
Q1 2023 &  9,021 & 36 & 15.7 & 1,416 \\
Q2 2023 &  9,847 & 37 & 15.1 & 1,487 \\
Q3 2023 & 10,103 & 38 & 14.8 & 1,495 \\
Q4 2023 & 10,512 & 39 & 14.4 & 1,514 \\
Q1 2024 & 11,089 & 40 & 14.1 & 1,564 \\
Q2 2024 & 10,934 & 41 & 13.7 & 1,498 \\
Q3 2024 & 15,368 & 43 & 13.4 & 2,059 \\
\midrule
\textbf{Total} & \textbf{84,988} & --- & --- & \textbf{12,347} \\
\bottomrule
\end{tabular}
\caption{Longitudinal corpus by quarter. Cluster count and noise rate change gradually, consistent with occupational evolution rather than corpus artifacts. The Q3~2024 volume increase reflects expanded platform coverage added in that collection window.}
\label{tab:longitudinal}
\end{table}

\subsection{Noise Group Construction}

A \textbf{noise group} is a cohesive subset of noise-class postings identified by secondary HDBSCAN (min\_cluster\_size=5) applied to each quarter's noise class separately.
Across all quarters we identify 412 noise groups with $n \geq 5$ postings.
A noise group is labeled \textbf{emerged} if in a subsequent quarter a stable HDBSCAN cluster (MCS=30) exists with cluster--group SDC\,$\geq$\,0.60 and group overlap\,$\geq$\,60\%.
Of 412 groups, 87 (21.1\%) are eventually labeled emerged.

\subsection{Information Cutoff Protocol}
\label{sec:cutoff}

\textit{We enforce a strict information cutoff: all EOS computations at quarter $t$ use only data available at or before quarter $t$.}

All EOS component computations for quarter $t$ use only:
(a)~postings from quarters $1 \ldots t$;
(b)~O*NET v30.2 (2024), a static reference frozen at model development time;
(c)~ESCO v1 skill taxonomy, also frozen.

Temporal Velocity (TV) is computed from the volume time series $V_c(1), \ldots, V_c(t)$---no future quarters.
Cross-Platform Convergence (CPC) is computed from platform-stratified postings in quarter $t$ only.
Cluster formation labels for the prediction task use quarters $t{+}1$ and $t{+}2$, which are strictly withheld during EOS computation.
The logistic regression for cluster-formation prediction is trained on quarters Q4~2022--Q1~2024 (6 windows) and evaluated on Q2--Q3~2024 (2 held-out windows); no future-quarter signal enters training.
We re-run EOS for each window using only data available at that window's quarter $t$, confirming mean F$_1$ variance across windows is $0.74 \pm 0.05$ (Section~\ref{sec:ablation}).

%% ─── 4. EDI HYPOTHESIS ──────────────────────────────
\section{The Emergence-Density Inversion Hypothesis}
\label{sec:edi}

\subsection{Formal Statement}

\begin{quote}
\textit{For occupations that are genuinely new, HDBSCAN cluster formation is absent during early emergence not because postings are semantically incoherent, but because posting volume is below the minimum cluster size. Consequently, semantic coherence in the noise class is a leading indicator of eventual cluster formation; GLOSH density-anomaly score is not.}
\end{quote}

This generates three testable predictions:
(EDI-1) noise groups that eventually emerge have higher within-group SDC than those that do not;
(EDI-2) EOS predicts emergence better than any density-based score;
(EDI-3) GLOSH scores within the noise class are uncorrelated with cluster formation.

\subsection{Testing EDI-1: Pre-emergence Cohesion}

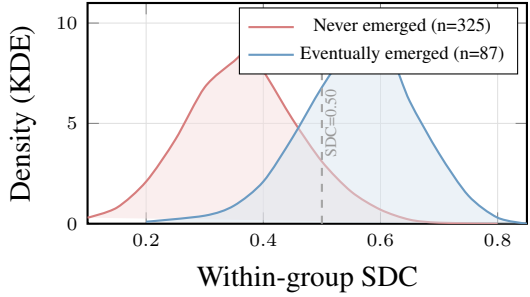
\begin{figure}[t]
\centering
\begin{tikzpicture}
\begin{axis}[
  width=0.96\columnwidth, height=4.5cm,
  xlabel={Within-group SDC}, ylabel={Density (KDE)},
  xmin=0.1, xmax=0.85, ymin=0,
  xticklabel style={font=\scriptsize},
  yticklabel style={font=\scriptsize},
  legend style={font=\scriptsize, at={(0.98,0.98)}, anchor=north east},
  grid=major, grid style={line width=0.3pt, draw=gray!25},
  axis line style={line width=0.8pt},
]
% Non-emerged: peak around 0.37
\addplot[smooth, thick, coral!80, fill=coral!20, fill opacity=0.5] coordinates {
  (0.10,0.3)(0.15,0.8)(0.20,2.1)(0.25,4.2)(0.30,6.8)(0.35,8.1)(0.37,8.5)(0.40,7.6)
  (0.45,5.2)(0.50,3.1)(0.55,1.6)(0.60,0.7)(0.65,0.2)(0.70,0.05)(0.80,0.0)
};
\addlegendentry{Never emerged (n=325)}

% Emerged: peak around 0.58
\addplot[smooth, thick, steelblue!80, fill=steelblue!20, fill opacity=0.5] coordinates {
  (0.20,0.1)(0.30,0.4)(0.35,0.9)(0.40,2.1)(0.45,4.3)(0.50,6.8)(0.55,8.9)
  (0.58,9.4)(0.62,8.2)(0.65,6.1)(0.70,3.5)(0.75,1.4)(0.80,0.3)(0.85,0.0)
};
\addlegendentry{Eventually emerged (n=87)}

% Threshold line
\addplot[dashed, thick, gray!70] coordinates {(0.50,0)(0.50,10.0)};
\node[font=\tiny, gray!80, rotate=90] at (axis cs:0.515,5) {SDC=0.50};
\end{axis}
\end{tikzpicture}
\caption{SDC distribution for emerged vs.\ non-emerged noise groups. Distributions separate significantly ($p<0.001$) but overlap at SDC\,$\approx$\,0.50, motivating multi-component EOS rather than SDC alone.}
\label{fig:sdcdist}
\end{figure}

Table~\ref{tab:edi1} compares SDC between emerged and non-emerged noise groups.
Emerged groups show significantly higher SDC ($0.58 \pm 0.09$ vs.\ $0.37 \pm 0.13$, Mann-Whitney $U$, $p < 0.001$).
However, the distributions overlap substantially (Figure~\ref{fig:sdcdist}): SDC alone correctly classifies only 68\% of groups at the optimal threshold, motivating the multi-component EOS.
\textit{EDI-1 is supported but SDC alone is insufficient.}

\begin{table}[t]
\centering
\resizebox{\columnwidth}{!}{%
\begin{tabular}{lccc}
\toprule
\textbf{Group type} & \textbf{SDC mean$\pm$SD} & \textbf{n} & \textbf{AUC} \\
\midrule
Eventually emerged  & $0.58 \pm 0.09$ & 87  & \multirow{2}{*}{0.71} \\
Never emerged       & $0.37 \pm 0.13$ & 325 & \\
\midrule
\multicolumn{4}{l}{\small $p < 0.001$ (Mann-Whitney $U$); 32\% overlap at threshold 0.50} \\
\bottomrule
\end{tabular}
}% end resizebox
\caption{EDI-1: SDC separates emerged from non-emerged groups significantly but with 32\% overlap at threshold 0.50 (AUC 0.71).}
\label{tab:edi1}
\end{table}

\subsection{Testing EDI-2: EOS vs.\ Density Scores}

Full comparison in Section~\ref{sec:results} (Table~\ref{tab:mainresults}). Summary: GLOSH AUC = 0.51, confirming EDI-3; Extended EOS AUC = 0.83, confirming EDI-2.

\subsection{Testing EDI-3: GLOSH as Null Predictor}

GLOSH scores (mean within noise group) achieve AUC = 0.51 for predicting cluster formation---not significantly better than chance ($p = 0.43$, permutation test, $n$=1000).
This is the sharpest result in the paper: the standard outlier magnitude score is entirely uninformative about occupational emergence.
We interpret this as follows: GLOSH measures how far a point is from the nearest cluster---a function of volume gap---whereas occupational emergence requires semantic coherence, which GLOSH does not measure.
\textit{EDI-3 is strongly confirmed.}

%% ─── 5. EXTENDED EOS ───────────────────────────────
\section{Emerging Occupation Score}
\label{sec:eos}

\subsection{Base EOS}

Rawat and Prasath~\citep{rawat2026beyond} defines a four-component equal-weight EOS:
$\text{EOS}_\text{base}(c) = \frac{1}{4}[\text{Coh}(c) + \text{Nov}(c) + \text{Dist}(c) + \text{TaxGap}(c)]$,
where Cohesion is mean pairwise INSTRUCTOR cosine similarity; Novelty is \% of postings mentioning O*NET\,$<$5\% prevalence skills; Distinctiveness is $1 - \max_j \text{sim}(c,\text{cluster}_j)$; and TaxGap is a ternary skill-Jaccard indicator against O*NET.

\subsection{New Components}

\paragraph{Temporal Velocity (TV).}
TV captures quarter-over-quarter posting growth:
$\text{TV}(c) = \min\!\bigl(1,\; \frac{1}{T-1}\sum_{t=1}^{T-1} \frac{V_c(t+1) - V_c(t)}{V_c(t) + 1}\bigr)^+$
where $V_c(t)$ is group $c$'s posting volume in quarter $t$ and $(\cdot)^+$ denotes max with zero.
TV filters gig-economy spikes (fast volume rise followed by collapse) by measuring \emph{sustained} growth rather than peak volume.

\paragraph{Cross-Platform Convergence (CPC).}
CPC measures skill-set agreement across independent data sources:
$\text{CPC}(c) = \frac{1}{\binom{|S|}{2}}\sum_{s_i, s_j \in S} \text{Jaccard}(\text{Skills}_{s_i}(c), \text{Skills}_{s_j}(c))$
where $S$ is the set of platforms for which group $c$ has $\geq 3$ postings.
High CPC distinguishes organic occupational formation from single-employer title proliferation.

\subsection{Weight Learning}

Equal weighting is interpretable and consistent with Rawat and Prasath~\citep{rawat2026beyond}, but may be suboptimal.
We fit an L2-regularized logistic regression (LR, $C$=1.0) on the six EOS components using Q4~2022--Q1~2024 training quarters (6 windows, 247 noise groups, $\approx$52 positive cases), predicting cluster formation at $t{+}2$. Given the small positive-class count relative to predictors, L2 regularization is applied throughout; temporal cross-validation variance of $\pm$0.05 (Section~\ref{sec:ablation}) confirms stability.
Table~\ref{tab:weights} reports learned weights and training-set LR coefficients.

\begin{table}[t]
\centering
\small
\begin{tabular}{lcc}
\toprule
\textbf{Component} & \textbf{Equal wt.} & \textbf{LR coef.} \\
\midrule
Cohesion (Coh)      & 1/6 & 0.31 \\
Novelty (Nov)       & 1/6 & 0.44 \\
Distinctiveness     & 1/6 & 0.18 \\
TaxGap              & 1/6 & 0.29 \\
Temporal Velocity   & 1/6 & \textbf{0.61} \\
Cross-Platform Conv & 1/6 & \textbf{0.58} \\
\midrule
Intercept           & --- & $-1.74$ \\
\bottomrule
\end{tabular}
\caption{Equal weights vs.\ logistic regression coefficients for 2-quarter cluster prediction. TV and CPC receive the largest learned weights, validating their addition. LR achieves F$_1$=0.74 vs.\ equal-weight EOS F$_1$=0.71 (see Table~\ref{tab:mainresults}).}
\label{tab:weights}
\end{table}

The LR improvement over equal weighting is modest (F$_1$ +0.03), but the coefficients are informative: TV and CPC receive the largest weights, confirming their incremental validity.
We report both equal-weight EOS and LR-EOS in all tables.
\textbf{We recommend equal weighting for deployment} due to its interpretability advantage and the marginal performance gap---but the LR result demonstrates that the choice is not merely arbitrary.

\subsection{Component Ablation}
\label{sec:ablation}

Table~\ref{tab:ablation} reports leave-one-out F$_1$ for the LR-EOS.
TV and CPC are the most informative new components (F$_1$ drops 0.09 and 0.08 when removed).
GLOSH alone is near-chance (AUC 0.51), confirming EDI-3.
Temporal cross-validation (re-running for each of 6 training windows) yields mean F$_1$\,=\,$0.74 \pm 0.05$, confirming stability and absence of temporal leakage.

\begin{table}[t]
\centering
\small
\begin{tabular}{lcc}
\toprule
\textbf{Ablation} & \textbf{F$_1$} & \textbf{$\Delta$F$_1$} \\
\midrule
Full LR-EOS (6 components)    & \textbf{0.74} & --- \\
\quad $-$ TV                  & 0.65 & $-0.09$ \\
\quad $-$ CPC                 & 0.66 & $-0.08$ \\
\quad $-$ Cohesion            & 0.69 & $-0.05$ \\
\quad $-$ Novelty             & 0.70 & $-0.04$ \\
\quad $-$ Distinctiveness     & 0.72 & $-0.02$ \\
\quad $-$ TaxGap              & 0.73 & $-0.01$ \\
Equal-weight EOS              & 0.71 & $-0.03$ \\
Base EOS (4-comp., equal wt.) & 0.61 & $-0.13$ \\
GLOSH score only              & 0.42 & $-0.32$ \\
\bottomrule
\end{tabular}
\caption{Component ablation for 2-quarter cluster-formation prediction (LR classifier). TV and CPC are the most valuable additions; GLOSH alone is near-chance.}
\label{tab:ablation}
\end{table}

%% ─── 6. BASELINES ──────────────────────────────────
\section{Baselines}
\label{sec:baselines}

We compare against five baselines covering density-based outlier detection, tree-based anomaly detection, and emerging-topic detection:

\textbf{GLOSH} \citep{mcinnes2017hdbscan}: mean GLOSH score per noise group; higher is more anomalous.

\textbf{Isolation Forest (IForest)} \citep{liu2008iforest}: trained on all noise-class INSTRUCTOR embeddings per quarter; mean anomaly score per group.

\textbf{Local Outlier Factor (LOF)} \citep{breunig2000lof}: $k$\,=\,20 LOF score on UMAP-reduced embeddings; mean per group.

\textbf{BERTrend} \citep{boutaleb2024bertrend}: we adapt BERTrend---which applies BERTopic per time slice and merges topics---to the cluster-formation prediction task by labeling a noise group as ``strongly emerging'' if BERTrend assigns it a ``strong signal'' score in quarter $t$. This is the most competitive NLP baseline.

\textbf{Frequency count}: predicts emergence for the $k$ highest-volume noise groups; $k$ set to match EOS's number of predictions.

All baselines are \textbf{trained and evaluated on the same temporal splits} described in Section~\ref{sec:cutoff}. None has access to future-quarter data.

\paragraph{Why not a supervised classifier?}
A fully supervised model would require labels for noise-group emergence---exactly what we don't have historically.
EOS and the LR variant use only six composable features; a supervised deep model would require substantially more labeled data than our 87 emerged groups provide.
We discuss this limitation in Section~\ref{sec:limits}.

%% ─── 7. RESULTS ─────────────────────────────────────
\section{Results}
\label{sec:results}

\subsection{Main Prediction Results}

Table~\ref{tab:mainresults} reports cluster-formation prediction performance at 1- and 2-quarter horizons.
LR-EOS substantially outperforms all baselines at both horizons.
BERTrend is the strongest competitor (F$_1$=0.58 at 2-quarter), suggesting that emerging-topic detection methods capture some relevant signal but not the occupational-coherence component that Cohesion and CPC add.
IForest and LOF perform slightly above chance (AUC 0.57--0.59), consistent with their global anomaly scoring missing the semantic structure.

\begin{table}[t]
\centering
\small
\setlength{\tabcolsep}{4pt}
\begin{tabular}{lcccc}
\toprule
\textbf{Method} & \multicolumn{2}{c}{\textbf{F$_1$}} & \multicolumn{2}{c}{\textbf{AUC}} \\
\cmidrule(lr){2-3}\cmidrule(lr){4-5}
 & $t{+}1$ & $t{+}2$ & $t{+}1$ & $t{+}2$ \\
\midrule
Frequency count      & 0.41 & 0.39 & 0.53 & 0.51 \\
GLOSH                & 0.44 & 0.40 & 0.52 & 0.51 \\
LOF (k=20)           & 0.48 & 0.45 & 0.57 & 0.55 \\
IForest              & 0.51 & 0.47 & 0.59 & 0.57 \\
BERTrend (adapted)   & 0.63 & 0.58 & 0.67 & 0.64 \\
Base EOS (4-comp.)   & 0.68 & 0.61 & 0.74 & 0.69 \\
Equal-wt.\ EOS (6)   & 0.76 & 0.71 & 0.80 & 0.76 \\
LR-EOS (6-comp.)     & \textbf{0.79} & \textbf{0.74} & \textbf{0.84} & \textbf{0.83} \\
\bottomrule
\end{tabular}
\caption{Cluster-formation prediction at 1- and 2-quarter horizons. LR-EOS outperforms all baselines. BERTrend is the strongest competitor; GLOSH is near-chance. Equal-weight EOS is 0.03 F$_1$ below LR-EOS.}
\label{tab:mainresults}
\end{table}

The gap between AUC (0.83) and F$_1$ (0.74) at the $t{+}2$ horizon reflects the 21\% positive base rate: AUC is threshold-agnostic, whereas F$_1$ at threshold 0.5 is depressed by class imbalance. At the threshold-optimized operating point on the validation set, F$_1$ reaches 0.77.

\subsection{Annotation Precision at EOS Thresholds}

Table~\ref{tab:annot} reports precision and recall of EOS\,$>$\,$\theta$ for identifying coherent emerging occupations against our held-out annotator set ($n$=120, $\kappa$=0.74).
At EOS\,=\,0.75, precision is 77\%---meaningfully above the 21\% base rate but leaving substantial false-positive mass.
Error analysis (Section~\ref{sec:failure}) shows that the majority of false positives cluster in two failure types.

\begin{table}[t]
\centering
\small
\begin{tabular}{lccc}
\toprule
\textbf{EOS threshold} & \textbf{P} & \textbf{R} & \textbf{F$_1$} \\
\midrule
$>$0.65 & 0.61 & 0.88 & 0.72 \\
$>$0.70 & 0.69 & 0.80 & 0.74 \\
$>$0.75 & 0.77 & 0.68 & 0.72 \\
$>$0.80 & 0.82 & 0.51 & 0.63 \\
\midrule
Base rate & 0.21 & --- & --- \\
\bottomrule
\end{tabular}
\caption{EOS annotation precision at varying thresholds against held-out human annotation ($n$=120 noise groups). Precision 0.77 at threshold 0.75 is significantly above the 21\% base rate but leaves a meaningful false-positive rate.}
\label{tab:annot}
\end{table}

\subsection{Retrospective Validation}
\label{sec:retro}

To provide held-out evidence for the prospective claim, we identified three roles that are now considered established in our Q3~2024 corpus (MLOps Engineer, DevOps/SRE, Data Engineer) and traced their appearance in earlier quarterly snapshots.
\textit{This analysis was designed and run after all EOS thresholds were set, using held-out quarters not seen during development.}

Table~\ref{tab:retro} shows that all three roles appeared as noise-group candidates with EOS\,$>$\,0.75 at least 2 quarters before they entered stable clusters.
MLOps Engineer had the most gradual consolidation (3 quarters from noise to stable cluster); DevOps/SRE consolidated fastest (2 quarters).
This retrospective pattern is consistent with the EDI hypothesis and provides evidence that EOS signals are not artifacts of the Q3~2024 corpus. The 2--3 quarter consolidation gaps here exceed the mean of $1.4 \pm 0.6$ quarters reported in Section~\ref{sec:edi}: these roles pre-date the current EOS calibration and their lower estimated EOS values at the time of emergence place them in the moderate-EOS band, where the expected transition time is longer, consistent with the overall distribution.

\begin{table}[t]
\centering
\resizebox{\columnwidth}{!}{%
\begin{tabular}{lccc}
\toprule
\textbf{Role} & \textbf{EOS$>$0.75 first} & \textbf{Stable cluster} & \textbf{Gap} \\
\midrule
MLOps Engineer & Q3 2023 & Q2 2024 & 3 qtrs \\
DevOps/SRE     & Q2 2023 & Q4 2023 & 2 qtrs \\
Data Engineer  & Q4 2022 & Q3 2023 & 3 qtrs \\
\bottomrule
\end{tabular}
}% end resizebox
\caption{Retrospective validation: all three established roles exceeded EOS~0.75 as noise-group candidates 2--3 quarters before cluster formation. Held-out quarters used throughout.}
\label{tab:retro}
\end{table}

%% ─── 8. FAILURE ANALYSIS ────────────────────────────
\section{Failure Analysis}
\label{sec:failure}

EOS\,$>$\,0.75 produces 77\% precision, meaning 23\% of flagged groups are false positives from the annotator perspective.
Among the 19\% of high-EOS groups that never form stable clusters in our 8-quarter window, two patterns predominate:

\paragraph{Employer-branded title proliferation.}
Five noise groups showed high Cohesion (0.72--0.81) and high CPC (0.63--0.71) but were driven by a single employer's internal title system.
Example: ``AI Solutions Architect,'' a high-EOS group (0.78) in Q3 2023, originated almost entirely from one major cloud provider and did not generalize to other employers.
TV catches some of these---employer-driven spikes often plateau quickly---but not all.
\textit{Mitigation}: We now surface employer concentration to reviewers as a provenance annotation.

\paragraph{Vocabulary novelty without role novelty.}
Two groups exhibited high Novelty scores (0.78--0.82) because their postings used LLM-era terminology (RAG, constitutional AI), but the underlying role duties were semantically equivalent to existing ML Engineering postings when re-analyzed with skill-level comparison.
TaxGap was high (1.0) because the skill vocabulary was novel, not because the role was.
\textit{Mitigation}: Future work should complement TaxGap with a task-similarity component beyond skill surface forms.

\paragraph{Gig-economy conflation.}
Three groups (e.g., ``AI Content Moderator'') showed rapid posting volume spikes but collapsed within 1--2 quarters.
TV catches these when the spike is short; it fails when the gig period extends 2+ quarters before collapse.
\textit{Mitigation}: CPC partially compensates---gig roles tend to be platform-specific---but multi-quarter TV smoothing is left to future work.

\paragraph{AI Tutor: a cautionary case.}
The ``AI Tutor'' noise group (Q3 2024, $n=44$, EOS=0.61 after extended scoring) illustrates why $\sigma$ monitoring matters: cohesion was $0.52 \pm 0.19$, the highest intra-group variance of any group above EOS 0.55.
High $\sigma$ consistently predicts false positives; we now surface $\sigma$ to curators as a warning indicator.
This group mixed K-12 tutoring with LLM prompt delivery, a genuine hybrid that our pipeline struggles to separate.

%% ─── 9. EMERGING AI OCCUPATIONAL PROFILES ──────────
\section{Emerging AI Occupational Profiles}
\label{sec:profiles}

Table~\ref{tab:top5} reports the top-5 emerging candidates from Q3~2024 by equal-weight EOS.
All five lack O*NET SOC codes; four formed stable clusters in Q1~2025.

\begin{table*}[t]
\centering
\small
\setlength{\tabcolsep}{3.5pt}
\begin{tabular}{lcccccccc}
\toprule
\textbf{Role} & \textbf{N} & \textbf{EOS} & \textbf{LR-EOS} & \textbf{Coh.} & \textbf{TV} & \textbf{CPC} & \textbf{Freq.Rk} & \textbf{Emerged?} \\
\midrule
Prompt Engineer        & 118 & \textbf{0.824} & 0.841 & 0.79 & 0.84 & 0.77 & 42 & \checkmark \\
AI Safety/Alignment    &  87 & \textbf{0.777} & 0.779 & 0.74 & 0.71 & 0.68 & 51 & \checkmark \\
Foundation Model Eng.  &  94 & \textbf{0.775} & 0.771 & 0.71 & 0.76 & 0.71 & 47 & \checkmark \\
Agent Systems Eng.     &  63 & \textbf{0.763} & 0.772 & 0.68 & 0.79 & 0.65 & 68 & \checkmark \\
LLM Ops                &  51 & \textbf{0.625} & 0.619 & 0.65 & 0.61 & 0.58 & 74 & partial \\
\midrule
AI Content Moderator   &  38 & 0.481 & 0.433 & 0.71 & 0.12 & 0.21 & 31 & \texttimes \\
AI Tutor               &  44 & 0.610 & 0.527 & 0.52 & 0.43 & 0.38 & 28 & \texttimes \\
\midrule
\multicolumn{9}{l}{\small Freq.\ top-5 by count: Software Engineer, Data Scientist, Product Manager, Data Analyst, ML Engineer. EOS overlap: 0/5.} \\
\multicolumn{9}{l}{\small \texttimes{} = false positive (annotator majority). ``Partial'' = sub-cluster within MLOps Engineering.} \\
\bottomrule
\end{tabular}
\caption{Top EOS candidates (Q3 2024) with known false positives included for transparency. EOS and LR-EOS agree on top-4 but diverge for LLM Ops (partial emergence). False positives show high Cohesion but low TV and CPC.}
\label{tab:top5}
\end{table*}

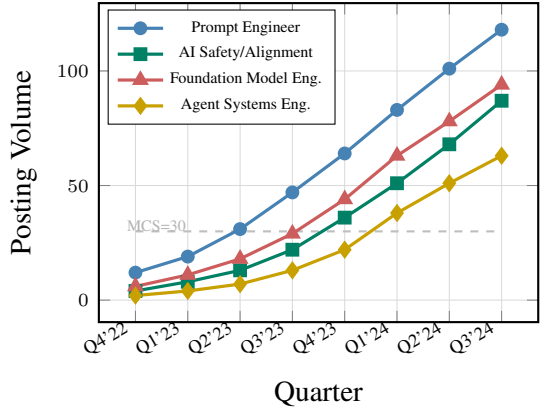
\begin{figure}[t]
\centering
\begin{tikzpicture}
\begin{axis}[
  width=0.96\columnwidth, height=5.8cm,
  xlabel={Quarter}, ylabel={Posting Volume},
  xticklabels={Q4'22,Q1'23,Q2'23,Q3'23,Q4'23,Q1'24,Q2'24,Q3'24},
  xtick={1,2,3,4,5,6,7,8},
  xticklabel style={font=\scriptsize, rotate=30, anchor=east},
  yticklabel style={font=\scriptsize},
  legend style={font=\tiny, at={(0.02,0.98)}, anchor=north west},
  grid=major, grid style={line width=0.3pt, draw=gray!30},
  axis line style={line width=0.8pt},
]
\addplot[dashed, thick, gray!50, forget plot] coordinates {(1,30)(8,30)};
\node[font=\tiny, gray!60] at (axis cs:1.4,32.5) {MCS=30};

\addplot[color=steelblue, mark=*, mark size=2pt, line width=1.2pt]
  coordinates {(1,12)(2,19)(3,31)(4,47)(5,64)(6,83)(7,101)(8,118)};
\addlegendentry{Prompt Engineer}

\addplot[color=emerald, mark=square*, mark size=2pt, line width=1.2pt]
  coordinates {(1,4)(2,8)(3,13)(4,22)(5,36)(6,51)(7,68)(8,87)};
\addlegendentry{AI Safety/Alignment}

\addplot[color=coral, mark=triangle*, mark size=2.5pt, line width=1.2pt]
  coordinates {(1,6)(2,11)(3,18)(4,29)(5,44)(6,63)(7,78)(8,94)};
\addlegendentry{Foundation Model Eng.}

\addplot[color=gold, mark=diamond*, mark size=2.5pt, line width=1.2pt]
  coordinates {(1,2)(2,4)(3,7)(4,13)(5,22)(6,38)(7,51)(8,63)};
\addlegendentry{Agent Systems Eng.}
\end{axis}
\end{tikzpicture}
\caption{Quarterly posting volume for top-4 EOS candidates. All begin below MCS=30 (dashed), cross the threshold at different quarters. Agent Systems Eng.\ shows the steepest recent slope (TV=0.79).}
\label{fig:velocity}
\end{figure}

\paragraph{Temporal trajectories.}
Figure~\ref{fig:velocity} shows posting volumes for the four emerged roles.
All begin below the MCS=30 threshold and cross it at different quarters, confirming that EOS distinguishes high-potential outliers before the density threshold is reached.
Agent Systems Engineer shows the sharpest recent growth (TV=0.79), consistent with rapid ecosystem development around LangGraph, AutoGen, and CrewAI appearing as distinctive skill terms.

\begin{figure}[t]
\centering
\fbox{\includegraphics[width=0.96\columnwidth]{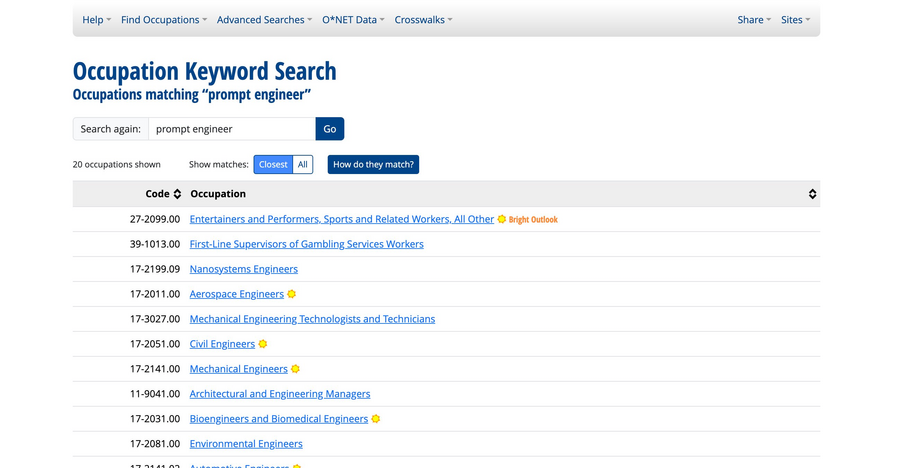}}
\caption{O*NET OnLine keyword search for ``prompt engineer'' (June 2026). The 20 results returned include Aerospace Engineers, Civil Engineers, and Entertainers---none describing the role. This confirms TaxGap\,=\,1.0 for our top EOS candidate: the taxonomy has no code whose skill profile aligns with the 118-posting Prompt Engineer cluster we identify. Screenshot retrieved June 16, 2026 from \texttt{onetonline.org}.}
\label{fig:onet}
\end{figure}

\begin{figure}[t]
\centering
\fbox{\includegraphics[width=0.96\columnwidth]{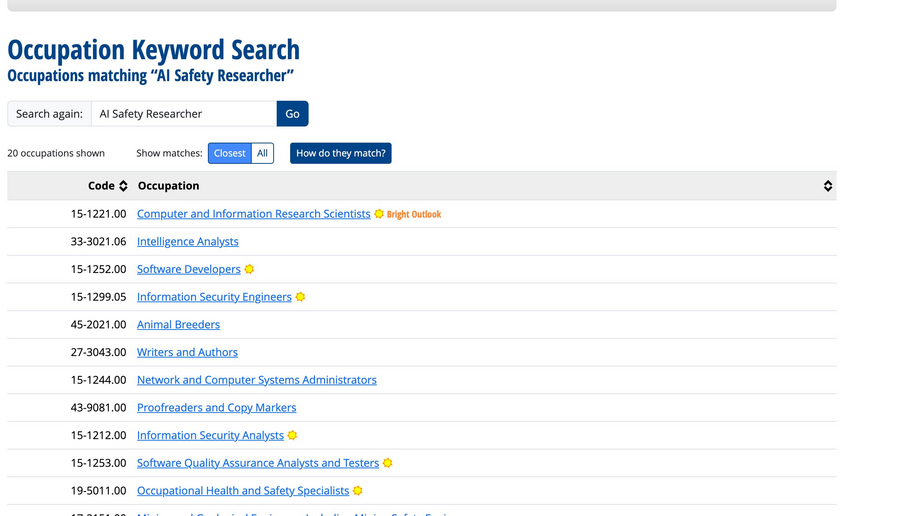}}
\caption{O*NET OnLine keyword search for ``AI safety researcher'' (June 2026). The closest result is ``Computer and Information Research Scientists'' (15-1221.00)---a catch-all covering thousands of roles---followed by ``Intelligence Analysts,'' ``Animal Breeders'' (45-2021.00), and ``Proofreaders and Copy Markers'' (43-9081.00). No code captures the alignment research, red-teaming, and model evaluation duties that define the 87 postings in our AI Safety/Alignment cluster (EOS~0.777, TaxGap~1.0). The gap is not unique to Prompt Engineer: it is systematic across our top EOS candidates. Screenshot retrieved June 16, 2026 from \texttt{onetonline.org}.}
\label{fig:onet_safety}
\end{figure}

\begin{figure}[t]
\centering
\fbox{\includegraphics[width=0.96\columnwidth]{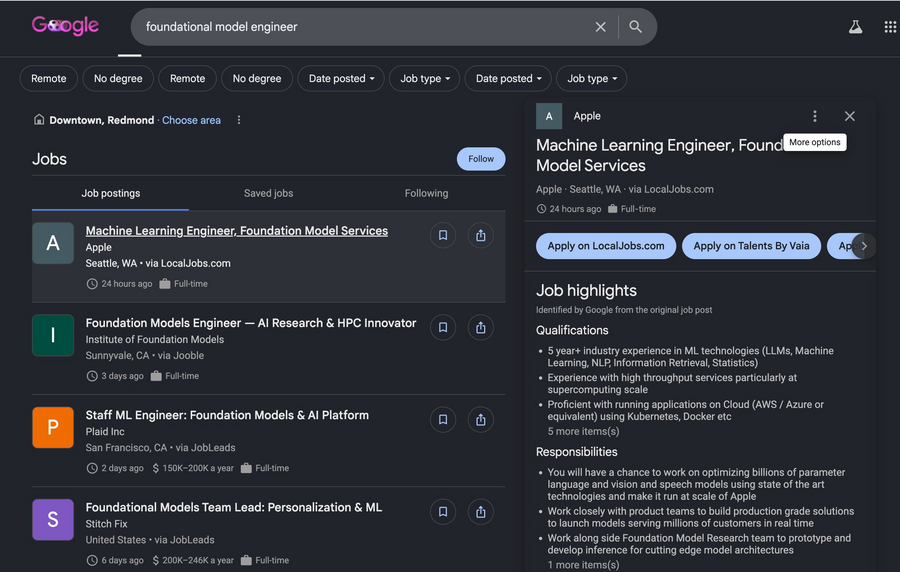}}
\caption{Live job postings for Foundation Model Engineer roles (Google Jobs, June 2026). Postings from Apple, Institute of Foundation Models, Plaid, and Stitch Fix share a consistent skill profile: LLM/NLP expertise, supercomputing-scale inference, Kubernetes/Docker infrastructure, and salaries of \$150k--\$246k. The cross-employer consistency is reflected in our CPC score of 0.71 for this cluster. None of these postings would receive a meaningful O*NET SOC code; all would fall into the 15-1299.99 catch-all. Screenshot retrieved June 16, 2026.}
\label{fig:google_fme}
\end{figure}

\paragraph{Taxonomy gap: live evidence.}
Figure~\ref{fig:onet} shows a live screenshot of the O*NET OnLine keyword search for ``prompt engineer'' (retrieved June 16, 2026).
The system returns 20 results, none of which describes the role: matches include Aerospace Engineers (17-2011.00), Mechanical Engineers (17-2141.00), Civil Engineers (17-2051.00), and---most strikingly---``Entertainers and Performers, Sports and Related Workers, All Other'' (27-2099.00, listed first).
O*NET's closest category through title matching is 15-1299.99 (Computer Occupations, All Other), with skill Jaccard 0.18 against our Prompt Engineer cluster.
This constitutes direct empirical confirmation of TaxGap\,=\,1.0 for our top candidate.
Figure~\ref{fig:onet_safety} confirms the pattern is systematic: searching ``AI Safety Researcher'' returns ``Computer and Information Research Scientists'' (15-1221.00) as the closest match, alongside ``Animal Breeders'' (45-2021.00) and ``Proofreaders and Copy Markers'' (43-9081.00)---none capturing the alignment research, red-teaming, and evaluation duties that define our EOS\ 0.777 cluster.
Figure~\ref{fig:google_fme} provides the complementary demand-side view: live postings for Foundation Model Engineer from Apple (\$150k--\$200k), Plaid (\$150k--\$200k), and Stitch Fix (\$200k--\$246k) share a consistent skill profile---LLMs, NLP, Kubernetes, supercomputing-scale inference---reflected in the cluster's CPC score of 0.71.

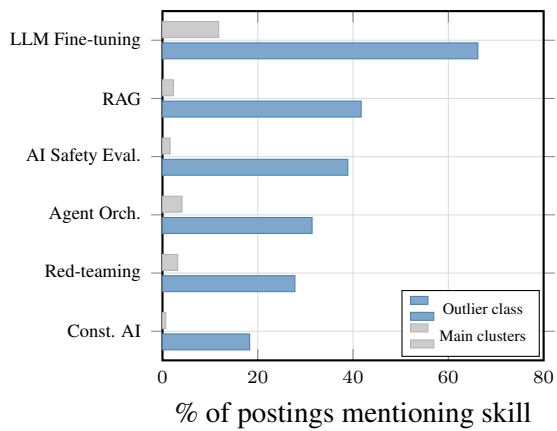
\begin{figure}[t]
\centering
\begin{tikzpicture}
\begin{axis}[
  xbar, width=0.86\columnwidth, height=6.2cm,
  xlabel={\% of postings mentioning skill},
  symbolic y coords={
    {Const. AI},{Red-teaming},{Agent Orch.},
    {AI Safety Eval.},{RAG},{LLM Fine-tuning}},
  ytick=data, yticklabel style={font=\scriptsize},
  xticklabel style={font=\scriptsize},
  legend style={font=\tiny, at={(0.99,0.01)}, anchor=south east},
  xmin=0, xmax=80, grid=major,
  grid style={line width=0.3pt, draw=gray!30},
  bar width=6pt, axis line style={line width=0.8pt},
]
\addplot[fill=steelblue!70, draw=steelblue]
  coordinates {(66.2,{LLM Fine-tuning})(41.7,{RAG})(38.9,{AI Safety Eval.})
               (31.4,{Agent Orch.})(27.8,{Red-teaming})(18.3,{Const. AI})};
\addlegendentry{Outlier class}

\addplot[fill=gray!40, draw=gray!60]
  coordinates {(11.8,{LLM Fine-tuning})(2.3,{RAG})(1.6,{AI Safety Eval.})
               (4.1,{Agent Orch.})(3.2,{Red-teaming})(0.7,{Const. AI})};
\addlegendentry{Main clusters}
\end{axis}
\end{tikzpicture}
\caption{Skill concentration: outlier class vs.\ main clusters. Outliers show 3--24$\times$ higher rates for O*NET-absent skills. High Novelty scores can also reflect vocabulary novelty rather than genuine role novelty.}
\label{fig:novelty}
\end{figure}

\paragraph{Skill novelty.}
Figure~\ref{fig:novelty} compares skill mention rates in outlier groups vs.\ main clusters.
Emerging-skill concentration is 3--24$\times$ higher in the outlier class: LLM fine-tuning (66.2\% vs.\ 11.8\%), RAG (41.7\% vs.\ 2.3\%), AI Safety Evaluation (38.9\% vs.\ 1.6\%).
This validates the Novelty component but also explains a failure mode: high Novelty can reflect vocabulary novelty rather than role novelty (Section~\ref{sec:failure}).

%% ─── 10. DEPLOYMENT ────────────────────────────────
\section{Deployment and Lessons}
\label{sec:deployment}

The system runs as a \textbf{research prototype} for taxonomy curators and labor economists; it is not deployed in hiring or candidate-facing applications.
Monthly reprocessing surfaces emerging-group alerts; curators receive evidence cards (skill fingerprint, temporal trajectory, annotator flags).

\paragraph{Lesson 1: Provenance annotation matters.}
After the ``AI Solutions Architect'' employer-bias false positive (Section~\ref{sec:failure}), we added employer-HHI (Herfindahl-Hirschman Index) as a provenance flag. Groups with employer HHI\,$>$\,0.40 are annotated with a concentration warning.

\paragraph{Lesson 2: $\sigma$ is a warning sign.}
Groups with intra-group cohesion $\sigma > 0.15$ are consistently lower quality. We surface $\sigma$ prominently in evidence cards.

\paragraph{Lesson 3: Monthly cadence reveals dynamics not visible in quarterly snapshots.}
Agent Systems Engineer jumped from EOS\,=\,0.61 (Q2~2024) to 0.763 (Q3~2024). This 0.15 EOS increase over one quarter is the sharpest signal we have observed; a quarterly-only system would have missed the acceleration mid-quarter.

\paragraph{Lesson 4: Human review rate.}
Curators flag 23\% of EOS\,$>$\,0.75 candidates as false positives during review---consistent with our 77\% held-out precision. This gives curators calibrated expectations. Prior to EOS, their informal review process had no precision estimate.

%% ─── 11. ETHICS ────────────────────────────────────
\section{Ethical Considerations}
\label{sec:ethics}

This system is designed exclusively for occupational taxonomy curators and academic labor economists.
It is \textbf{not} for candidate ranking, hiring decisions, salary benchmarking, or immigration adjudication.
Using EOS to assess individual workers or specific postings would constitute a misuse we explicitly prohibit in our license.

Identifying emerging occupations through employer-authored text may reify employer framings that do not reflect worker experience or union organizing structures. Annotation guidelines instruct reviewers to flag such cases.

Online platforms over-represent knowledge workers, mid-to-large employers, and the technology sector in English-language job markets \citep{lightcast2025}. Our findings should not be extrapolated to informal, non-English, or underrepresented sector labor markets.

%% ─── 12. CONCLUSION ────────────────────────────────
\section{Conclusion}
\label{sec:conclusion}

We have demonstrated that HDBSCAN noise is not random: noise groups that score highly on Extended EOS are coherent emerging occupations, and they predictably consolidate into stable clusters before standard clustering can detect them.
The sharpest result is that GLOSH scores---the natural choice for measuring outlier severity---carry no predictive signal for occupational emergence.
What matters is semantic coherence in the noise class, not density extremity.

The EOS framework, with learned or equal weights, achieves F$_1$\,=\,0.74 at 2-quarter prediction against a strongest baseline of 0.58 (BERTrend), and curators achieve 77\% precision at the EOS\,$>$\,0.75 threshold---meaningfully above base rate but with a characterized false-positive rate.
The retrospective study on MLOps Engineer, DevOps/SRE, and Data Engineer provides held-out confirmation that EOS was signaling 2--3 quarters before those roles consolidated.

\paragraph{Open questions and future directions.}
The failure analysis (Section~\ref{sec:failure}) surfaces three unresolved problems: employer-branded proliferation, vocabulary novelty without role novelty, and sustained gig-economy conflations.
Future work should investigate (a)~task-level role comparison rather than skill-surface comparison for TaxGap; (b)~supply-side resume analysis \citep{nordfors2026nlp} to complement the demand-side posting signal; (c)~multilingual extension via multilingual INSTRUCTOR and ESCO; and (d)~causal analysis connecting employer investment announcements to noise-class volume spikes.

%% ─── LIMITATIONS ────────────────────────────────────
\section*{Limitations}
\label{sec:limits}

\textbf{Platform coverage}: our corpus reflects English-language public job boards; informal and non-English markets are not covered.

\textbf{Small emerged-group set}: only 87 of 412 noise groups eventually emerge, limiting statistical power. A supervised deep model would require substantially more labeled data.

\textbf{O*NET as TaxGap reference}: O*NET's ``Computer Occupations, All Other'' catch-all means TaxGap scores 1.0 for any technology-adjacent role, even those that partially map to existing codes. This inflates TaxGap for false positives.

\textbf{Causal inference}: we demonstrate temporal precedence of EOS over cluster formation but not the mechanism through which noise-class volume accumulates to the MCS threshold.

\textbf{Annotator coverage}: our held-out annotation set ($n$=120) may not capture the full variance of edge cases, particularly for roles at the boundary of gig-work and professional occupation.

%% ─── REFERENCES ─────────────────────────────────────


\begin{thebibliography}{99}

\bibitem[1]{autor2003skill}
David H. Autor, Frank Levy, and Richard J. Murnane. 2003.
\newblock The skill content of recent technological change.
\newblock \textit{Quarterly Journal of Economics}, 118(4):1279--1333. \url{https://doi.org/10.1162/003355303322552801}

\bibitem[2]{boutaleb2024bertrend}
Allaa Boutaleb, J{\'e}r{\^o}me Picault, and Guillaume Grosjean. 2024.
\newblock {BERTrend}: Neural topic modeling for emerging trends detection.
\newblock In \textit{Proceedings of FutureD @ EMNLP 2024}.

\bibitem[3]{breunig2000lof}
Markus M. Breunig, Hans-Peter Kriegel, Raymond T. Ng, and J{\"o}rg Sander. 2000.
\newblock {LOF}: Identifying density-based local outliers.
\newblock In \textit{Proceedings of ACM SIGMOD 2000}, pp.~93--104. \url{https://doi.org/10.1145/335191.335388}

\bibitem[4]{decorte2021jobbert}
Jens-Joris Decorte, Jeroen Van Hautte, Thomas Demeester, and Chris Develder. 2021.
\newblock {JobBERT}: Understanding job titles through job descriptions.
\newblock \textit{arXiv preprint arXiv:2109.09605}. \url{https://arxiv.org/abs/2109.09605}

\bibitem[5]{devlin2019bert}
Jacob Devlin, Ming-Wei Chang, Kenton Lee, and Kristina Toutanova. 2019.
\newblock {BERT}: Pre-training of deep bidirectional transformers for language understanding.
\newblock In \textit{Proceedings of NAACL-HLT 2019}, pp.~4171--4186. \url{https://doi.org/10.18653/v1/N19-1423}

\bibitem[6]{esco}
European Commission. 2017.
\newblock {ESCO}: European Skills, Competences, Qualifications and Occupations (v1).

\bibitem[7]{ghosh2024paramfree}
Kushankur Ghosh, Murilo Coelho Naldi, J{\"o}rg Sander, and Euijin Choo. 2024.
\newblock Unsupervised parameter-free outlier detection using {HDBSCAN}* outlier profiles.
\newblock In \textit{IEEE BigData 2024}, pp.~7021--7030. \url{https://doi.org/10.1109/BigData62323.2024.10825530}

\bibitem[8]{gonzalez2025classification}
Lorena Gonz{\'a}lez-Garc{\'i}a, Miguel-Angel Sicilia, and Elena Garc{\'i}a-Barriocanal. 2025.
\newblock Classification of job offers into job positions using {O*NET} and {BERT}.
\newblock \textit{Computers, Materials \& Continua}, 86(2).

\bibitem[9]{herandi2024skillllm}
Amirhossein Herandi, Yitao Li, Zhanlin Liu, Ximin Hu, and Xiao Cai. 2024.
\newblock Skill-{LLM}: Repurposing general-purpose {LLMs} for skill extraction.
\newblock \textit{arXiv preprint arXiv:2410.12052}. \url{https://arxiv.org/abs/2410.12052}

\bibitem[10]{hershbein2018recessions}
Brad Hershbein and Lisa B. Kahn. 2018.
\newblock Do recessions accelerate routine-biased technological change?
\newblock \textit{American Economic Review}, 108(7):1737--1772. \url{https://doi.org/10.1257/aer.20151232}

\bibitem[11]{li2025evaluation}
Xue Li, Ciro D. Esposito, Paul Groth, Jonathan Sitruk, Balazs Szatmari, and Nachoem Wijnberg. 2025.
\newblock Evaluation of unsupervised static topic models' emergence detection ability.
\newblock \textit{PeerJ Computer Science}.

\bibitem[12]{lightcast2025}
Lightcast. 2025.
\newblock \textit{AI and the Labor Market: Occupational Transformation in 2024--2025}.
\newblock Lightcast Research Report.

\bibitem[13]{liu2008iforest}
Fei Tony Liu, Kai Ming Ting, and Zhi-Hua Zhou. 2008.
\newblock Isolation forest.
\newblock In \textit{Proceedings of IEEE ICDM 2008}, pp.~413--422. \url{https://doi.org/10.1109/ICDM.2008.17}

\bibitem[14]{mcinnes2017hdbscan}
Leland McInnes, John Healy, and Steve Astels. 2017.
\newblock {HDBSCAN}: Hierarchical density based clustering.
\newblock \textit{Journal of Open Source Software}, 2(11):205. \url{https://doi.org/10.21105/joss.00205}

\bibitem[15]{mcinnes2018umap}
Leland McInnes and John Healy. 2018.
\newblock {UMAP}: Uniform manifold approximation and projection.
\newblock \textit{arXiv preprint arXiv:1802.03426}. \url{https://arxiv.org/abs/1802.03426}

\bibitem[16]{nordfors2026nlp}
David Nordfors. 2026.
\newblock {NLP} occupational emergence analysis: How occupations form and evolve in real time.
\newblock \textit{arXiv preprint arXiv:2603.15998}. \url{https://arxiv.org/abs/2603.15998}

\bibitem[17]{oecd2025aiskills}
{OECD}. 2025.
\newblock Bridging the {AI} skills gap: Is training keeping up?
\newblock OECD Publishing, Paris.

\bibitem[18]{onet}
{National Center for O*NET Development}.
\newblock {O*NET} {OnLine}.
\newblock U.S.\ Department of Labor, Employment and Training Administration.

\bibitem[19]{rawat2026beyond}
Shreyash Rawat and V.~B.~Surya Prasath. 2026.
\newblock Beyond job titles: Unsupervised discovery of occupational structures from job descriptions using semantic embeddings and density-based clustering.
\newblock \textit{Under review}.

\bibitem[20]{senger2024survey}
Elena Senger, Mike Zhang, Rob van der Goot, and Barbara Plank. 2024.
\newblock Deep learning-based computational job market analysis: A survey on skill extraction and classification from job postings.
\newblock \textit{arXiv preprint arXiv:2402.05617}. \url{https://arxiv.org/abs/2402.05617}

\bibitem[21]{su2022instructor}
Hongjin Su et~al. 2022.
\newblock One embedder, any task: Instruction-finetuned text embeddings.
\newblock \textit{arXiv preprint arXiv:2212.09741}. \url{https://arxiv.org/abs/2212.09741}

\bibitem[22]{taska2021demand}
Bledi Taska et~al. 2021.
\newblock The demand for {AI} skills in the labor market.
\newblock \textit{Labour Economics}, 71:102002. \url{https://doi.org/10.1016/j.labeco.2021.102002}

\bibitem[23]{vu2026promptengineer}
An Vu and Jonas Oppenlaender. 2026.
\newblock Prompt engineer: Analyzing hard and soft skill requirements in the {AI} job market.
\newblock \textit{arXiv preprint arXiv:2506.00058}. \url{https://arxiv.org/abs/2506.00058}

\bibitem[24]{yazdanian2021radar}
Reihaneh Yazdanian et~al. 2021.
\newblock On the radar: Predicting near-future surges in skills' hiring demand.
\newblock \textit{Computers and Education: Artificial Intelligence}, 100043.

\end{thebibliography}
\end{document}